\pdfoutput=1

\documentclass[11pt]{article}

\usepackage{acl}

\usepackage{times}
\usepackage{latexsym}
\usepackage{graphicx}
\usepackage{amsmath} 
\usepackage{subcaption}
\usepackage{comment}

\usepackage{booktabs}       
\usepackage{amsfonts}       
\usepackage{nicefrac}       
\usepackage{microtype}      
\usepackage{xcolor}         
\usepackage{amssymb}
\usepackage{multirow}
\usepackage{array}
\usepackage{url}

\usepackage[T1]{fontenc}

\usepackage[utf8]{inputenc}

\usepackage{microtype}

\usepackage{inconsolata}

\title{Rebuilding ROME : Resolving Model Collapse during \\Sequential Model Editing}

\author{$\text{Akshat Gupta}^1$, $\text{Sidharth Baskaran}^2$, $\text{Gopala Anumanchipalli}^1$\\
$^1\text{UC Berkeley}$, $^2\text{Automorphic Inc.}$\\
  \texttt{akshat.gupta@berkeley.edu, sid@automorphic.ai}}

\begin{document}
\maketitle
\begin{abstract}
Recent work using Rank-One Model Editing (ROME), a popular model editing method, has shown that there are certain facts that the algorithm is unable to edit without breaking the model. Such edits have previously been called disabling edits \cite{akshat-catastrophic}. These disabling edits cause immediate model collapse and limits the use of ROME for sequential editing. In this paper, we show that disabling edits are an artifact of irregularities in the implementation of ROME. With this paper, we provide a more stable implementation ROME, which we call r-ROME and show that model collapse is no longer observed when making large scale sequential edits with r-ROME, while further improving generalization and locality of model editing compared to the original implementation of ROME.
\end{abstract}

\section{Introduction}
Large language models (LLMs) are expensive to train and the knowledge contained in these models gets obsolete with time. Model editing or knowledge editing \citep{editing-survey} has recently come out as a popular method to update knowledge in large language models (LLMs). In this paper, we focus on one popular parameter-modifying model editing methods called ROME (Rank-One Model Editing) \cite{ROME}. ROME is not only one of the most popular model editing algorithms, but is also widely used in unlearning \cite{can-sensitive-info} and model interpretability \citep{geva-patchscope, geva-dissecting-factual-recall} literature.


While a lot of model editing approaches perform well when making singular edits, editing multiple facts in a model still remains a challenge for parameter-modifying model editing methods. One way to make multiple edits to the same model is through \textbf{sequential editing} \cite{editing-survey} - where we make a series of single edits to a model by modifying the parameters of the model after every edit. Recent works have started studying the effects of sequential editing and found that ROME \cite{ROME} was prone to a sudden model collapse by a single edit \citep{akshat-catastrophic, disabling-butterfly, disabling-wilke}. This effect was first observed in \citet{akshat-catastrophic} during sequential editing. The collapse included complete loss of downstream performance, inability to recall previously editing facts and loss of the ability to even get edited. Such facts were named \textbf{disabling edits} by \citet{akshat-catastrophic} and were later independently observed by \citet{disabling-butterfly, disabling-wilke}. 


\begin{figure}
    \centering
    \includegraphics[width=0.8\linewidth]{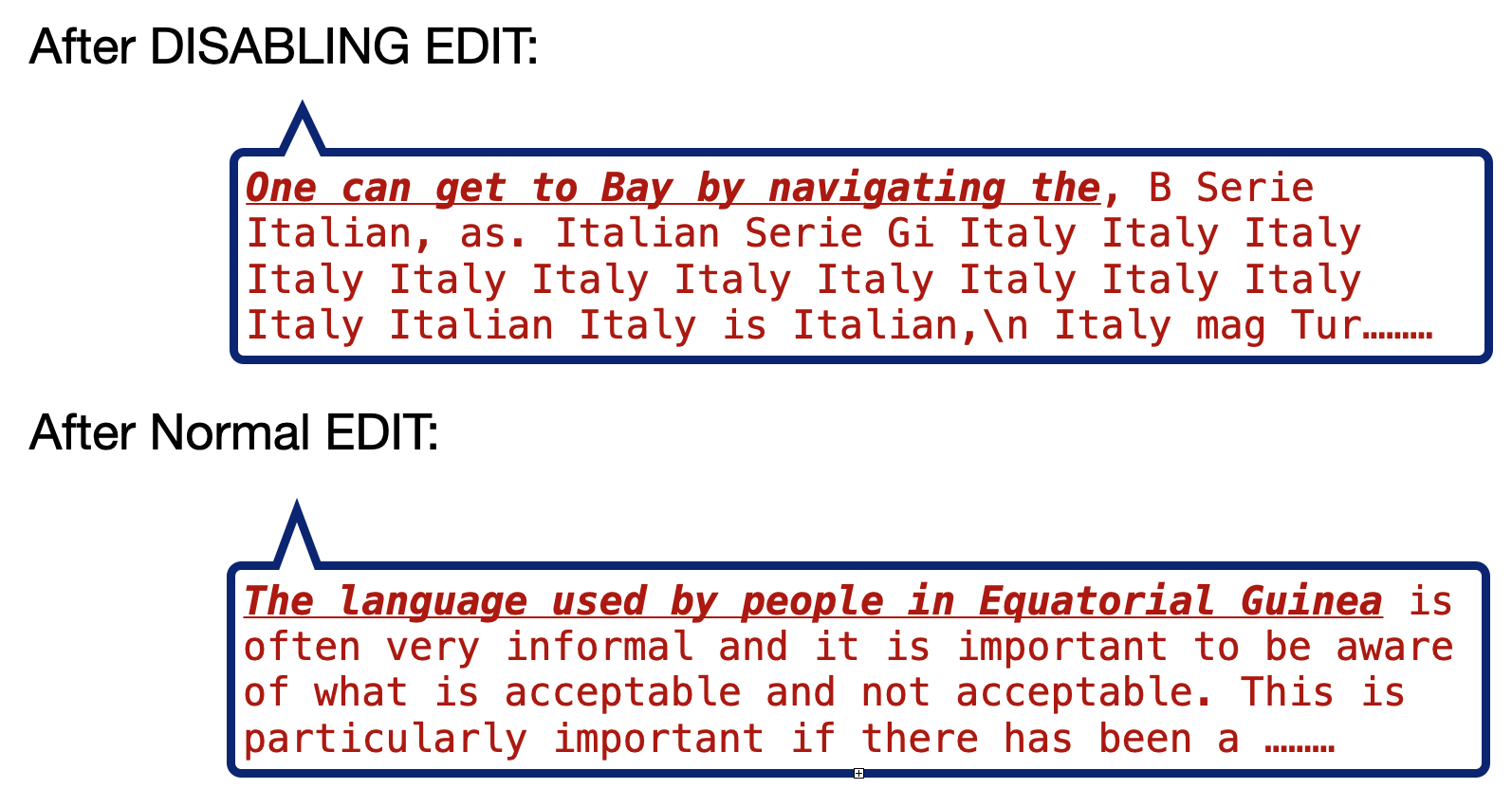}
    \caption{A typical generation example after a disabling edit is compared to a normal model edit using ROME. The bold and underlined part in the text is input prompt.}
    \label{fig:generation-example}
\end{figure}

Disabling edits are detrimental for knowledge editing at scale. While a gradual model degradation is expected as we make sequential edits to a model \cite{akshat-catastrophic}, disabling edits lead to a sudden model collapse irrespective of when the disabling fact is edited, making sequential editing impossible. An example of this can be seen in Figure \ref{fig:gptj-original-downstream}, where instead of allowing gradual model degradation when doing sequential editing like in Figure \ref{fig:sequential-r-rome-fixed-gptj}, the presence of disabling edits lead to a sudden and immediate model collapse.


\begin{table*}
    \vskip 0.05in
    \centering 
    \scriptsize
    \setlength\tabcolsep{0pt}
    \setlength\extrarowheight{1pt}
    \begin{tabular*}{\textwidth}{@{\extracolsep{\fill\centering}}*{9}{c}}
        \toprule 
        \multirow{2}{*}[-0.3em]{\textsc{Dataset}} & 
            \multirow{2}{*}[-0.3em]{\textsc{Implementation}} &
            \multicolumn{2}{c}{Efficacy} & 
            \multicolumn{2}{c}{Generalization} & 
            \multicolumn{2}{c}{Locality} & 
            \multicolumn{1}{c}{Score}\\ 
        \addlinespace[0.125em] \cline{3-9} \addlinespace[0.25em]
        &  & ES $\uparrow$ & EM $\uparrow$ &
             PS $\uparrow$ & PM $\uparrow$ &
             NS $\uparrow$ & NM $\uparrow$ &
            S $\uparrow$\\
        \midrule 
        \multirow{2}{*}[-0em]{CF} & \textsc{Original} & $99.92$ & $99.68$ & $96.29$ & $71.58$ & $75.8$ & $10.25$ &  $89.32$\\
        & r\textsc{-ROME}& $99.74$ & $97.79$ & $\mathbf{99.09}$ & $70.86$ & $\mathbf{80.62}$ & $26.0$ & $\mathbf{92.22}$ \\
        & p\textsc{-ROME}& $\mathbf{99.9}$ & $99.36$ & $97.04$ & $63.01$ & $80.0$ & $5.74$ & $91.42$ \\
        \midrule\bottomrule
    \end{tabular*}
        \caption{The above represents model editing results for 5000 singular model edits made on GPT-J-6B from the CounterFact dataset (non-sequential).}\label{table:comparison-independent}
    \vskip -0.0in
\end{table*}

In this paper, we aim to find the source of these disabling edits. We first introduce two metrics for identifying disabling edits - generation entropy and the norm of matrix update. We plot edits made by ROME along these two dimensions and show new ways of identifying disabling edits even when making singular edits. As we dig deeper into the optimization objectives and the codebase of ROME, we find that the disabling edits in ROME are a result of irregularities in the implementation of ROME, and not an artifact of the optimization objective. Specifically, disabling edits were caused due to the asymmetric usage of key-vectors in the update equation of ROME. With this paper, we share our new ROME code-base and invite researchers to use it for model editing. Our implementation of ROME, which we call r-ROME, can be found \href{https://github.com/scalable-model-editing/rebuilding-rome}{here}\footnote{\url{https://github.com/scalable-model-editing/rebuilding-rome}}.


\section{Background}\label{sec:background}

Facts are usually added in ROME using key-value format, where a key is the vector representation of a query-phrase and the value is the vector representation of the target object. For example, when adding a new fact - \textit{"The president of USA is John Cena"}, the query-phrase here is \textit{"The president of USA is"} and the target object is \textit{"John Cena"}.  The key-vector is defined by \citet{ROME} is the activation of the first linear layer in the MLP targeted by ROME:

\begin{align}
    k^{\left( l^* \right)}(x)= \sigma \left( W_{f c}^{\left( l^* \right)} \gamma\left( a_{[x], i}^{\left( l^* \right)} + h_{[x], i}^{\left( l^* -1 \right)} \right) + b_{fc}^{\left( l^* \right)} \right)
\end{align}

Editing in ROME is done using a pair of vectors - $(k_e, v_e)$ that represent a new fact being added. $k_e$, also called the key-vector is a vector representation of the query-phrase, and $v_e$, or the value-vector is the vector representation of the target object.
The weights of the specific layer being edited in ROME are updated from $W_0$ to $\hat{W}$ by inserting a new fact $(k_e, v_e)$ using the following equation:

\begin{equation}\label{eq:rome-update-equation}
\begin{aligned}  
    \hat{W} &= W_0 + \Delta \hspace{10pt} \\
    \text{where} \hspace{10pt} \Delta &= (v_e - W_0k_e) \frac{k_e^TC_0^{-1}}{k_e^TC_0^{-1}k_e} 
\end{aligned}
\end{equation}

where $\Delta$ is the update to the current weight matrix being edited such that the new fact $(k_e, v_e)$ gets incorporated. Additionally, each key-vector in $k_e$ is not just the representation of a single prompt. To enhance generalization, \citet{ROME, MEMIT} create the key-vector as an average representations over the query-phrase with random prefixes. This is done so that the represented key-vectors do not just represent one way to phrase the query-phrase and edits made using these representations can generalize over different paraphrases of the edited facts. The final key vector is found by averaging over $N$ random prefixes using the equation:

\begin{align}\label{eq:key-average}
    k_e = \frac{1}{N} \sum^{N}_{i = 1} k (x_i \oplus p)
\end{align}

Here $k(x_i \oplus p)$ represents the key-vector corresponding to a prefix $x_i$ being concatenated with the original query-phrase $p$. Examples of prefixes added in ROME can be seen in Table \ref{tab:prefixes}. In this paper, we will refer to the averaged prefix representation of keys with $k_e$, whereas when the representation just consists of the original prompt, we will depict that with a superscript as $k^o_e$. The following equation explicitly differentiates between the two mathematically: 

\begin{align}\label{eq:key-single}
    k^o_e = k(p)
\end{align}

\paragraph{Evaluating Model Editing.} Model editing is usually evaluated along three metrics - reliability, generalization and locality. Reliability represents if a fact was successfully added in a model and is measured using edit score (ES) and edit magnitude (EM) metrics. ES measures the portion of cases when an edited fact is more probable than the original fact post-editing, whereas EM measures the difference in the probability magnitudes of the edited and original facts. Generalization represents if the edited fact is recalled through paraphrases of the prompt used to edit the fact and is measured using paraphrase score (PS) and paraphrase magnitude defined similary as above for paraphases of the edited facts. Locality represents if editing of one fact affects other facts stored inside a model and is measured using neighborhood score (NS) and neighborhood magnitude (NM) on facts unrelated to the edited facts. The \textit{score} metric is the harmonic mean of ES, PS and NS. We follow standard model editing metrics proposed in the original ROME paper \citet{ROME}. We refer the reader to \citet{editing-survey, ROME} for a more comprehensive review of model editing metrics. 

Additionally, we also evaluated the model on downstream task performance as proposed by \citep{akshat-catastrophic}, which becomes especially important when making sequential edits to the same model. We evaluate the edited model on four tasks from the GLUE \cite{glue} benchmark - sentiment analysis (SST2), paraphrase detection (MRPC), natural language inference (NLI) and linguistic acceptability classification for doing downstream evaluation.

\begin{figure}
    \centering
    \begin{subfigure}{.24\textwidth}
        \centering
        \includegraphics[width=0.9\linewidth]{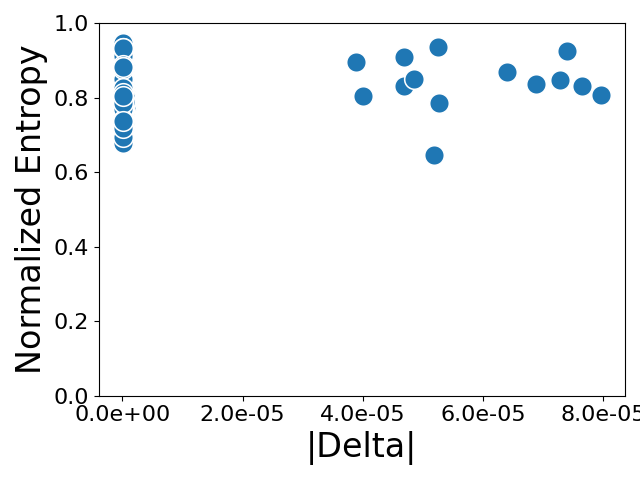}
        \caption{ROME}
        \label{fig:memit_gptj:edit_score}
    \end{subfigure}%
    \begin{subfigure}{.24\textwidth}
        \centering
        \includegraphics[width=0.9\linewidth]{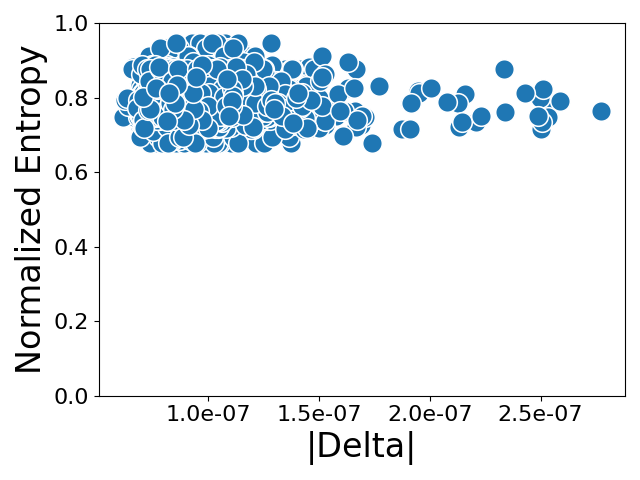}
        \caption{r-ROME}
        \label{fig:memit_gptj:forgetting}
    \end{subfigure}
    \caption{This figure shows the difference between the ROME and r-ROME updates on GPTJ (6B) for 5k individual edits. Our implementation shows much less potential disabling edits indicated by lower $|\Delta|$ values.}
    \label{fig:fixed-disabling}
\end{figure}

\begin{table*}
    \vskip 0.05in
    \centering 
    \scriptsize
    \setlength\tabcolsep{0pt}
    \setlength\extrarowheight{1pt}
    \begin{tabular*}{\textwidth}{@{\extracolsep{\fill\centering}}*{9}{c}}
        \toprule 
        \multirow{2}{*}[-0.3em]{\textsc{Dataset}} & 
            \multirow{2}{*}[-0.3em]{\textsc{Implementation}} &
            \multicolumn{2}{c}{Efficacy} & 
            \multicolumn{2}{c}{Generalization} & 
            \multicolumn{2}{c}{Locality} & 
            \multicolumn{1}{c}{Score}\\ 
        \addlinespace[0.125em] \cline{3-9} \addlinespace[0.25em]
        &  & ES $\uparrow$ & EM $\uparrow$ &
             PS $\uparrow$ & PM $\uparrow$ &
             NS $\uparrow$ & NM $\uparrow$ &
             S $\uparrow$\\
        \midrule 
        \multirow{2}{*}[-0em]{CF} & \textsc{Original} & $62.43$ & $11.23$ & $59.12$ & $7.49$ & $52.05$ & $-0.05$ & $57.53$\\
        & r\textsc{-ROME}& $97.92$ & $72.14$ & $96.23$ & $54.97$ & $59.52$ & $0.16$ & $80.20$ \\
        & p\textsc{-ROME}& $99.94$ & $95.31$ & $94.05$ & $55.22$ & $52.57$ & $-1.54$ & $75.64$ \\
        \midrule\bottomrule
    \end{tabular*}
       \caption{We find that our implementations (r-ROME \& and p-ROME) retains edit performance significantly more than the original implementation of ROME on standard model editing metrics for GPT-J-6B. We use the same 5k CounterFact examples from as Table \ref{table:comparison-independent} \textbf{sequentially}.}\label{table:comparison-sequential}
    \vskip -0.0in
\end{table*}

\section{Experiments}

\subsection{Properties of Disabling Edits}\label{sec:metrics}
Disabling edits \cite{akshat-catastrophic} are defined as singular knowledge edits that lead to sudden loss of ability to do downstream tasks or any kind of meaningful generation. \citet{akshat-catastrophic} also showed one way of identifying disabling edits was the unusually large norm of the update matrix. In other words, $|\Delta|$ in equation \ref{eq:rome-update-equation} was unusually higher when compared to normal edits.\footnote{$|\Delta| = \|\Delta\|_2/N$ is the L2 norm of the update matrix normalized by the number of elements in the update matrix.}

Figure \ref{fig:generation-example} shows a typical example of model collapse where the model constantly repeats a single word. The simplest metric to identify such a model collapse is to calculate the entropy over the probability distribution of vocabulary elements of text generated from the model. For this, a probability distribution is calculated over the vocabulary of a sample generation consisting of ten generations, and is normalized by the vocabulary size to remove the effect of the size of vocabulary. If the model collapses as shown in Figure \ref{fig:generation-example}, we expected the normalized entropy to be small and concentrated around a handful of words. 


The first set of experiments we do is to search for disabling edits. We do this by making singular model edits using ROME on GPT-J and GPT2-XL using the CounterFact dataset to replicate the conditions where disabling edits occurred in prior work. We measure the above mentioned metrics as shown in Figure \ref{fig:fixed-disabling}(a) for GPT-J. Similar patterns are observed for GPT2-XL and are shown in Figure \ref{fig:disabling_edits_gpt2xl} (appendix). When editing facts from the CounterFact dataset, we see two clusters forming. We find that certain edits have larger values of $|\Delta|$ for ROME, indicating the presence of disabling edits. 

\subsection{Fixing ROME}
\begin{figure}
    \centering
    \begin{subfigure}{.24\textwidth}
        \centering
        \includegraphics[width=0.9\linewidth]{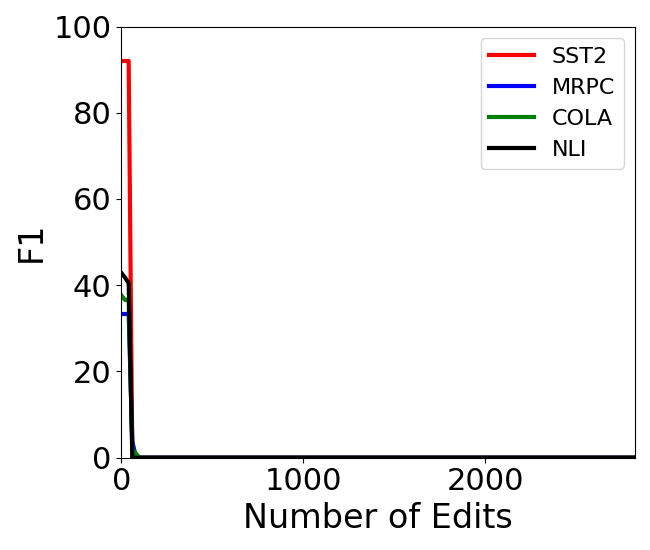}
        \caption{Downstream Evaluation}
        \label{fig:gptj-original-downstream}
    \end{subfigure}%
    \begin{subfigure}{.24\textwidth}
        \centering
        \includegraphics[width=0.9\linewidth]{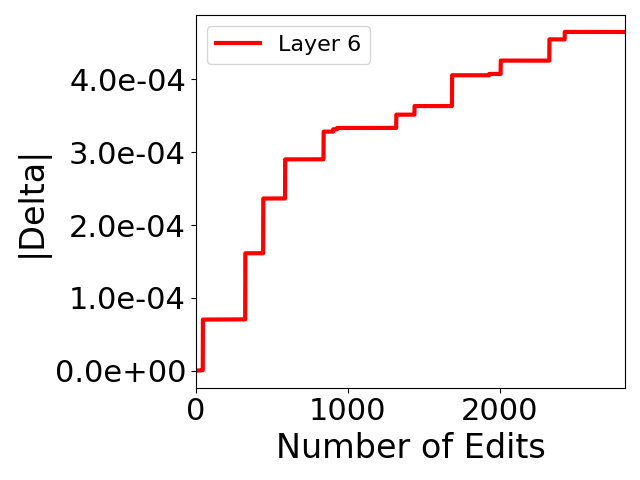}
        \caption{$|\Delta|$}
        \label{fig:gptj-original-distance}
    \end{subfigure}
    \caption{Sequential editing using original implementation of ROME on GPT-J (6B).}
    \label{fig:sequential-rome-original-gptj}
\end{figure}

After finding signals of disabling edits while making singular edits, we perform sequential editing with ROME. Every iteration of sequential editing with ROME leads to model collapse similar to Figure \ref{fig:sequential-rome-original-gptj}(a). This collapse occurs at random points during the editing process at one of the facts that clustered away in Figure \ref{fig:fixed-disabling}(a). 
After a long inquiry into the optimization objective of ROME, we found no reason for $|\Delta|$ of certain edits to be so large. We then turned to the implementation of ROME and found some interesting discrepancies. Although seemingly benign, these discrepancies eventually lead to disabling edits. The core reason behind disabling edits is that instead of implementing equation \ref{eq:rome-update-equation} as mentioned in the paper, the authors of ROME \cite{ROME} implement the following equation for $\Delta$:

\begin{equation}\label{eq:rome-imp}
\begin{aligned}  
 \hspace{10pt} \Delta_{imp} = (v_e - W_0 \mathbf{k^o_e}) \frac{k_e^TC_0^{-1}}{k_e^TC_0^{-1}\mathbf{k^o_e}} 
\end{aligned}
\end{equation}

where $\Delta_{imp}$ represents the actual implementation of $\Delta$ in the code by \citet{ROME}, with the difference highlighted in bold. The difference in implementation and original derivation of ROME is the use of two different types of key vectors. Rather than using key-vectors that average over prefix prompts or $k_e$ (eq \ref{eq:key-average}), the authors end up using $k^o_e$ (eq \ref{eq:key-single}) is certain places in the update equation. \textbf{We find that this asymmetry in usage of the key-vector causes disabling edits}.

To fix this issue, we create homogeneity in the usage of the key-vectors. We first use $k_e$ everywhere in the update equation, an implementation we refer to as \textbf{r-ROME}. This is the correct implementation of ROME as originally intended by the authors of \citet{ROME}. We then use keys generated using only the original prompts or $k^o_e$ homogeneously in the update equation, referred to as \textbf{p-ROME}. This also tests the hypothesis that using a key-vector averaged over random prefixes can create more generalizable edits.

The first evidence of removal of disabling edits can be seen in Figure \ref{fig:fixed-disabling}, where the $|\Delta|$ of the updates are orders of magnitude smaller for r-ROME when compared to the original implementation. The overall results for independent edits are shown in Table \ref{table:comparison-independent}. We find that edits made using r-ROME create more generalized edits at the slight expense of efficacy, resulting in a higher total edit score than the original implementation. p-ROME leads to increased efficacy and worse generalization resulting in a slightly lower edit score. This shows that homogeneity in using key-vectors is crucial in making model edits.


\begin{figure}
    \centering
    \begin{subfigure}{.24\textwidth}
        \centering
        \includegraphics[width=0.9\linewidth]{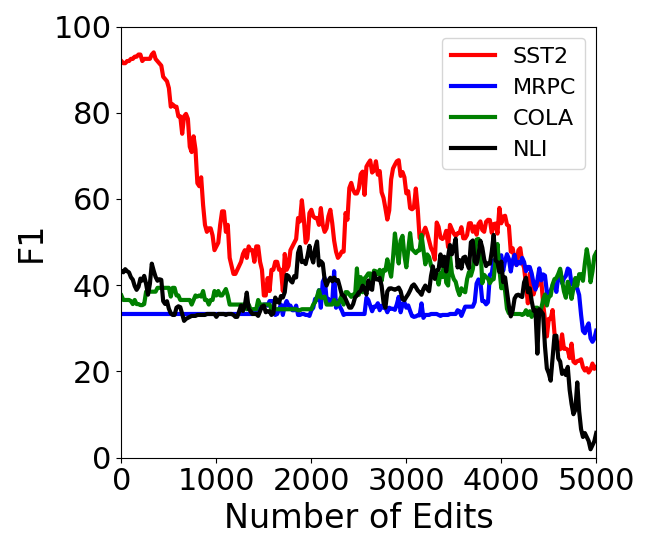}
        \caption{Downstream Evaluation}
        \label{fig:gptj-fixed-downstream}
    \end{subfigure}%
    \begin{subfigure}{.24\textwidth}
        \centering
        \includegraphics[width=0.9\linewidth]{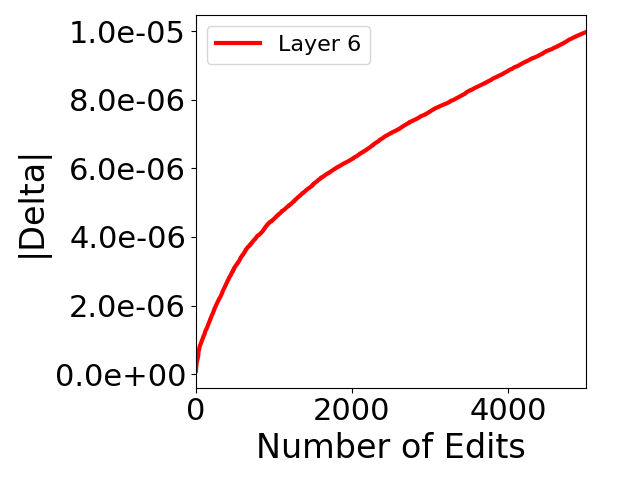}
        \caption{$|\Delta|$}
        \label{fig:gptj-fixed-downstream}
    \end{subfigure}
    \caption{Sequential editing with r-ROME on GPT-J.}
    \label{fig:sequential-r-rome-fixed-gptj}
\end{figure}

\subsection{Sequential Editing with r-ROME}

The final litmus test of r-ROME is to study its performance during large scale sequential editing. Figure \ref{fig:sequential-rome-original-gptj} shows a typical case of sequential editing using the original ROME code-base for GPT-J, where the presence of a disabling edit leads to large $|\Delta|$ and model collapse, as can be seen by an immediate loss of downstream performance in Figure \ref{fig:gptj-original-downstream}. With r-ROME (Figure \ref{fig:sequential-r-rome-fixed-gptj}), we see that $|\Delta|$ is orders of magnitude smaller and increases smoothly, which allows the model to maintain its general abilities and avoids model collapse. This enables large scale sequential model editing without loss of performance. The final model editing metrics after 5000 sequential edits for GPT-J are shown in Figure \ref{table:comparison-sequential}, with r-ROME significantly outperforming the original implementation of ROME.  Additional sequential editing results using p-ROME and GPT-XL can be found in section \ref{sec:sequential-gpt2xl}.

\section{Conclusion}
In this paper, we show that model edits made using the original implementation of ROME lead to unstable model edits eventually causing model collapse. Our re-implementations of ROME, called r-ROME (\href{https://github.com/scalable-model-editing/rebuilding-rome}{code}) prevents model collapse and leads to stable and scalable model edits, thus making sequential editing possible using ROME. We believe that such an improvement to the algorithm should be available to the widespread community, especially due to the potential impact and reach of ROME.

\section{Limitations}
The focus of our paper was to identify reasons behind model collapse when using ROME and to mitigate such effects. While r-ROME does that and enables sequential editing with ROME, downstream performance degradation and decreased stability (as observed from increasing $|\Delta|$) still occurs at scale. This is an inherent limitation of ROME that we do not overcome and is beyond the scope of this paper. 



\bibliography{custom}

\newpage
\appendix

\begin{table*}[h]
\footnotesize
\centering
\begin{tabular}{ll}
\toprule
\textbf{Original Prompt} & The President of the USA is  \\ 
\midrule
\textbf{Prefix Prompts} & The President of the USA is \\
 & Therefore, I like. The President of the USA is\\ 
 & He is a. The President of the USA is \\ 
 & Today is a sunnay day. The President of the USA is \\ 
 & On this day. The President of the USA is \\ 
\bottomrule
\end{tabular}
\caption{Table showing examples of random prefixes $x_i$ from \ref{eq:key-average} added to the original query-phrase.}\label{tab:prefixes}


\end{table*}

\section{Related Work}

Recent works \citep{akshat-catastrophic, disabling-butterfly, disabling-wilke} also observe the phenomenon of disabling edits as a result of performing sequential edits with parametric methods such as ROME and MEMIT \citep{MEMIT}. The sequential model editing task proves to be more difficult for parametric editing methods at scale due to model saturation and catastrophic forgetting. Non-parametric methods such as SERAC \citep{SERAC} bypass this limitation by maintaining an external edit memory that removes the distinction between batched (simultaneous) and sequential edits. We primarily focus on single edits via ROME in this paper, however, sequential editing can be combined with batching for better scalability \cite{emmet}.

\begin{figure}
        \centering
        \includegraphics[width=0.6\linewidth]{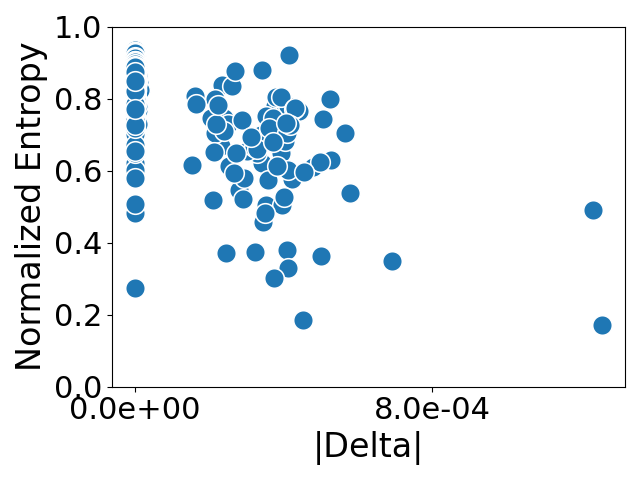}

    \caption{This figure shows distribution of edits along |Delta| and Normalized Entropy metric for edits using the original ROME implementation on CounterFact dataset for GPT2-XL.}
    \label{fig:disabling_edits_gpt2xl}
\end{figure}

\section{Additional Sequential Editing Experiments}\label{sec:sequential-gpt2xl}

\begin{table*}
    \vskip 0.05in
    \centering 
    \scriptsize
    \setlength\tabcolsep{0pt}
    \setlength\extrarowheight{1pt}
    \begin{tabular*}{\textwidth}{@{\extracolsep{\fill\centering}}*{9}{c}}
        \toprule 
        \multirow{2}{*}[-0.3em]{\textsc{Dataset}} & 
            \multirow{2}{*}[-0.3em]{\textsc{Implementation}} &
            \multicolumn{2}{c}{Efficacy} & 
            \multicolumn{2}{c}{Generalization} & 
            \multicolumn{2}{c}{Locality} & 
            \multicolumn{1}{c}{Score}\\ 
        \addlinespace[0.125em] \cline{3-9} \addlinespace[0.25em]
        &  & ES $\uparrow$ & EM $\uparrow$ &
             PS $\uparrow$ & PM $\uparrow$ &
             NS $\uparrow$ & NM $\uparrow$ &
            S $\uparrow$\\
        \midrule 
        \multirow{2}{*}[-0em]{CF} & \textsc{Original} & $99.94$ & $97.92$ & $96.38$ & $62.2$ & $75.8$ & $4.33$ &  $89.35$\\
        & r\textsc{-ROME}& $98.98$ & $93.35$ & $95.75$ & $59.65$ & $76.39$ & $4.63$ &  $89.18$ \\
        & p\textsc{-ROME}& $99.68$ & $97.68$ & $(88.67$ & $46.6$ & $76.28$ & $4.59$ &  $87.15$ \\
        \midrule\bottomrule
    \end{tabular*}
       \caption{Comparing the original implementation of ROME with  (r-ROME \& and p-ROME) for 5k non-sequential edits for GPT2-XL.}\label{table:comparison-gpt2xl}
    \vskip -0.0in
\end{table*}

The results for sequential edits on GPT-J are shown in Table \ref{table:comparison-sequential}. We indeed find that edits made using r-ROME create more generalized edits at the slight expense of efficacy as in \ref{table:comparison-independent} but downstream performance is retained at scale. The original implementation's downstream performance collapses almost immediately (\ref{fig:sequential-rome-original-gptj}). p-ROME surprisingly retains downstream performance better than r-ROME at the tail end of the sequential edits. We suspect this is related to the instability and noise the random prefixes induce: r-ROME n-gram entropies are more widely distributed than p-ROME (\ref{fig:fixed-disabling}).

\begin{figure}
    \centering
    \begin{subfigure}{.24\textwidth}
        \centering
        \includegraphics[width=\linewidth]{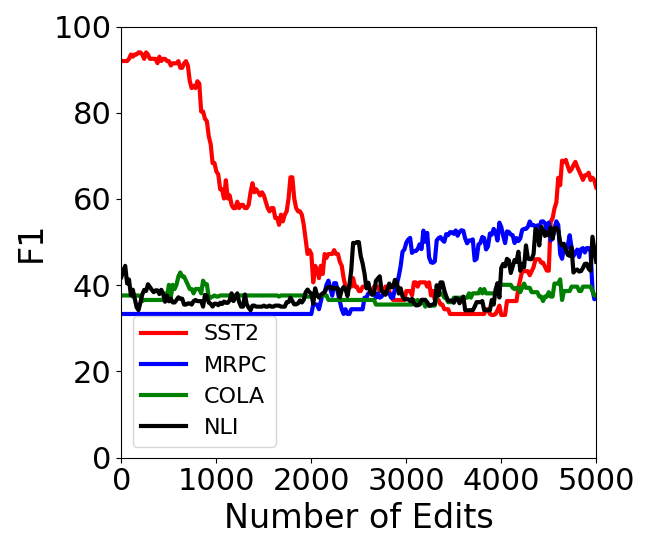}
        \caption{Downstream Evaluation}
        \label{fig:gptj-fixed-downstream}
    \end{subfigure}%
    \begin{subfigure}{.24\textwidth}
        \centering
        \includegraphics[width=\linewidth]{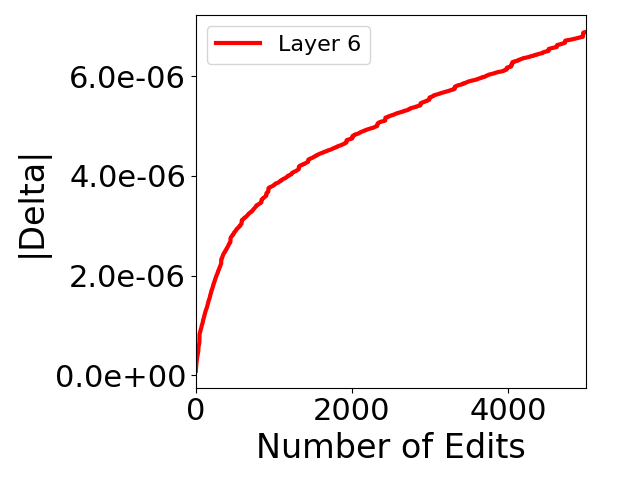}
        \caption{$|\Delta|$}
        \label{fig:gptj-fixed-downstream}
    \end{subfigure}
    \caption{Sequential editing with p-ROME on GPT-J (6B).}
    \label{fig:sequential-p-rome-fixed-gptj}
\end{figure}

We observe similar trends in the sequentuial editing scenario with GPT2-XL 1.5B as with GPT-J 6B. Notably, p-ROME performs worse in the downstream evaluations than r-ROME, we postulate that this is due to the poorer generalization ability of the smaller model; GPT-J's generalization abilities seem to bridge the downstream performance gap between r-ROME and p-ROME.

\begin{figure}[h]
    \centering
    \begin{subfigure}{.24\textwidth}
        \centering
        \includegraphics[width=\linewidth]{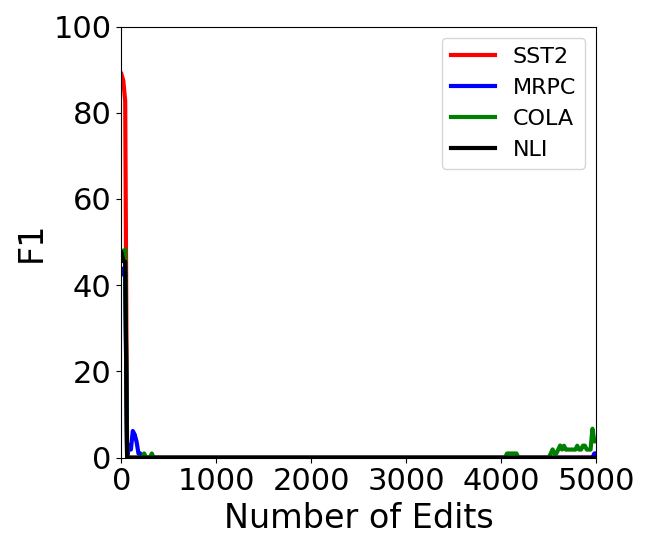}
        \caption{Downstream Evaluation}
        \label{fig:gpt2xl-original-downstream}
    \end{subfigure}%
    \begin{subfigure}{.24\textwidth}
        \centering
        \includegraphics[width=\linewidth]{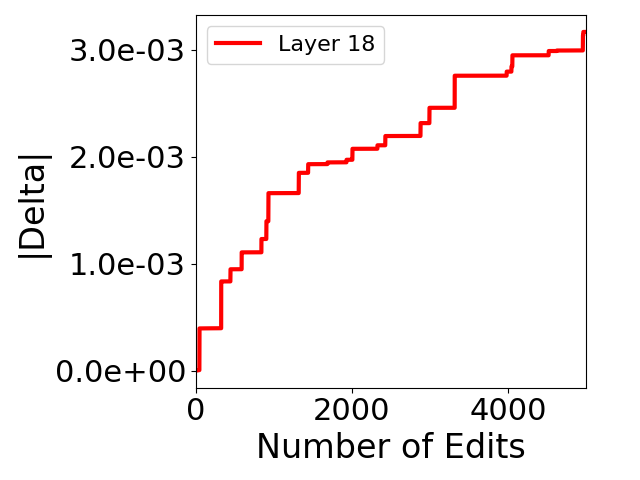}
        \caption{$|\Delta|$}
        \label{fig:gpt2xl-original-distance}
    \end{subfigure}
    \caption{Sequential editing using original implementation of ROME on GPT2-XL (1.5B) on the 5K CounterFact samples.}
    \label{fig:sequential-rome-original-gpt2xl}
\end{figure}
 
\begin{figure}[h]
    \centering
    \begin{subfigure}{.24\textwidth}
        \centering
        \includegraphics[width=\linewidth]{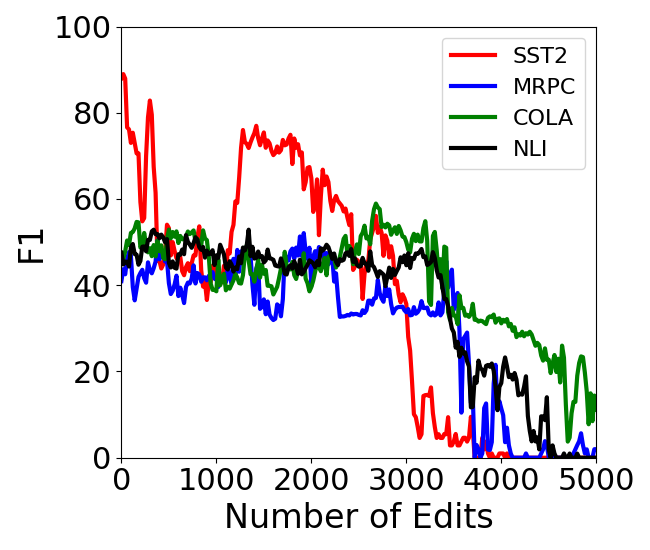}
        \caption{Downstream Evaluation}
        \label{fig:gpt2xl-fixed-downstream}
    \end{subfigure}%
    \begin{subfigure}{.24\textwidth}
        \centering
        \includegraphics[width=\linewidth]{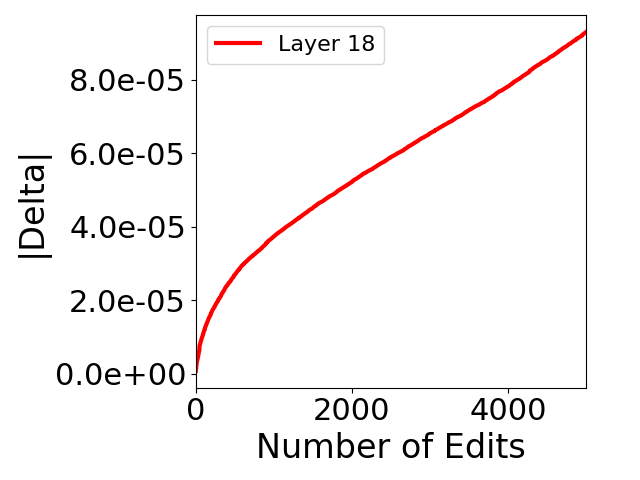}
        \caption{$|\Delta|$}
        \label{fig:gpt2xl-fixed-downstream}
    \end{subfigure}
    \caption{Sequential editing with r-ROME on GPT2-XL (1.5B) on the 5K CounterFact samples.}
    \label{fig:sequential-rome-fixed-gpt2xl}
\end{figure}

\begin{figure}[h]
    \centering
    \begin{subfigure}{.24\textwidth}
        \centering
        \includegraphics[width=\linewidth]{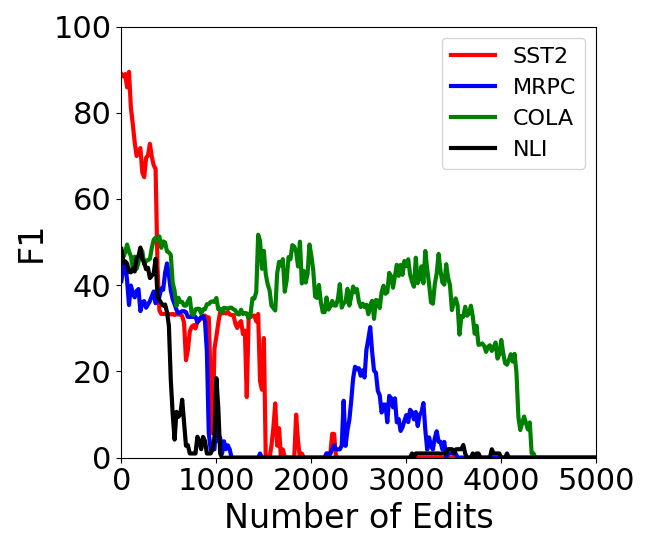}
        \caption{Downstream Evaluation}
        \label{fig:gpt2xl-fixed-downstream}
    \end{subfigure}%
    \begin{subfigure}{.24\textwidth}
        \centering
        \includegraphics[width=\linewidth]{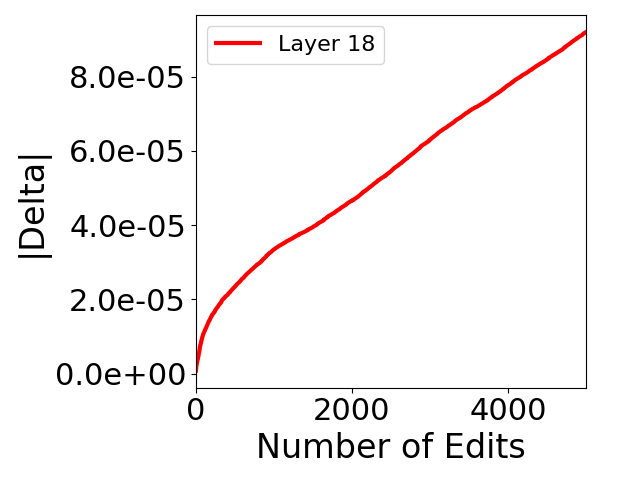}
        \caption{$|\Delta|$}
        \label{fig:gpt2xl-fixed-downstream}
    \end{subfigure}
    \caption{Sequential editing with p-ROME on GPT2-XL (1.5B) on the 5K CounterFact samples.}
    \label{fig:sequential-rome-fixed-gpt2xl}
\end{figure}

\end{document}